\documentclass{llncs}
\usepackage{url}

\usepackage{soul}

\usepackage[colorinlistoftodos]{todonotes}
\usepackage{verbatim}
\usepackage{subfig}
\usepackage{booktabs}
\usepackage{url}

\usepackage{xcolor}
\newcommand{\todoo}[1]{}
\renewcommand{\todoo}[1]{{\color{red} TODO: {#1}}}

\usepackage{amsmath}
\usepackage{graphicx}
\usepackage[colorinlistoftodos]{todonotes}

\title{EchoFusion: Tracking and Reconstruction of Objects in 4D Freehand Ultrasound Imaging without External Trackers}
\author{%
Bishesh Khanal\inst{1,2}  \and 
Alberto Gomez\inst{1}
Nicolas Toussaint\inst{1} \and 
Steven McDonagh\inst{2} \and
Veronika Zimmer\inst{1} \and
Emily Skelton\inst{1} \and 
Jacqueline Matthew\inst{1} \and
Daniel Grzech\inst{2} \and
Robert Wright\inst{1} \and
Chandni Gupta\inst{1} \and
Benjamin Hou\inst{2} \and
Daniel Rueckert\inst{2} \and
Julia A. Schnabel\inst{1} \and
Bernhard Kainz\inst{2}
}%
\institute{
School of Biomedical Engineering \& Imaging Sciences, King's College London, U.K.\\
\email{bishesh.khanal@kcl.ac.uk},\\ 
\and
Department of Computing, Imperial College London, U.K.
}

\begin{document}
\maketitle

\begin{abstract}
Ultrasound (US) is the most widely used fetal imaging technique.
However, US images have limited capture range, and suffer from view dependent artefacts such as acoustic shadows. Compounding of overlapping 3D US acquisitions into a high-resolution volume can extend the field of view and remove image artefacts, which is useful for retrospective analysis including population based studies.
However, such volume reconstructions require information about relative transformations between probe positions from which the individual volumes were acquired.
In prenatal US scans, the fetus can move independently from the mother, making external trackers such as electromagnetic or optical tracking unable to track the motion between probe position and the moving fetus.  
We provide a novel methodology for image-based tracking and volume reconstruction by combining recent advances in deep learning and simultaneous localisation and mapping (SLAM). 
Tracking semantics are established through the use of a Residual 3D U-Net and the output is fed to the SLAM algorithm.
As a proof of concept, experiments are conducted on US volumes taken from a whole body fetal phantom, and from the heads of real fetuses. For the fetal head segmentation, we also introduce a novel weak annotation approach to minimise the required manual effort for ground truth annotation.
We evaluate our method qualitatively, and quantitatively with respect to tissue discrimination accuracy and tracking robustness. 

\end{abstract}

\section{Introduction}

Ultrasound (US) is a very widely used medical imaging modality, well known for its portability, low cost, and high temporal resolution.
Although the most popular US imaging is 2D B-mode, 3D mode has become an attractive addition providing a larger field of view at an increased frame rate.
There is also growing interest in developing low cost 3D US probes~\cite{angiolini20171024}. 
While 2D mode images are usually of higher resolution, 3D mode has the ability to provide better context of the anatomy with smaller number of images.
Thus, 3D images could allow easier compounding and field of view extension to capture all the desired anatomy in a single compounded volume.

Volumetric compounding requires the relative transformation between individual volumes. 
This can be achieved using image registration if the offset is small and assumptions about the spatial arrangement of the volumes hold, e.g., when performing an imaging sweep at constant speed. 
For large offsets, or random views of a target volume, image registration alone is insufficient and external tracking such as electromagnetic or optical tracking has to be used to establish localisation coherence.
External tracking measures absolute transformations between a fiducial marker on the ultrasound probe and a calibrated world coordinate system. 
Moving targets within a patient cannot be tracked with fiducial markers, computer vision methods that rely on a direct line of sight, or by tracking the probe via external trackers. 

An ability to generate high quality compounded volumes of individual fetuses can be useful for retrospective analysis by experts who might not be available, e.g. in rural areas where the live scanning may be performed by non-experts.
High quality compounded volumes can also be important in creating US atlases of different fetal organs.
For example, it would be desirable to combine all possible views of the brain of single fetus to maximise the information obtained from individual fetal brains. 
In fetuses of late Gestational ages (GAs), acquiring images from all possible directions requires probe manipulation, incurring large rotation and translational motion.
Registration and tracking of images resulting from such constraint-free probe motions is typically highly challenging.
A motion-robust and hardware-lean image-based method to compound a large anatomical RoI in real-time is thus highly desired.

\textbf{Contribution:} 
We propose a novel approach to tackle the tracking problem during 3D fetal US examinations where an application-focused tissue discriminator, based on convolutional neural networks, is integrated into a simultaneous localisation and mapping (SLAM) formulation named EchoFusion. The proposed method yields relative transformations between subsequent volumes, surface reconstruction of the target anatomy, and reconstruction of a compounded volume at the same time.
We demonstrate the potential of the proposed approach with experiments for rigid whole body fetal phantom, and for free-hand 4D US covering the head region in real fetuses, without external tracking or a highly restrictive scanning protocol.
EchoFusion requires the fetal tissue discriminator to be accurate only in the fetal surface closest to the US probe, allowing the use of: i) challenging 4D fetal screening US images coming from a very wide range of views, and ii) weak annotations, enabling large training data at low cost.

\textbf{Related work: } 
Extending the FOV by compounding multiple 3D images has been in focus since a wide range of freehand ultrasound probes support 3D images with either matrix array transducers~\cite{Wygant2008} or mechanically steered linear arrays in plane fan mode~\cite{Fenster1996}. 
Tracking-based methods \cite{Dewi2015,Solberg2007} provide good initialisation for a variety of subsequent and task-specific registration methods but often need additional calibration to establish the transformation between object and tracking coordinate system~\cite{Blackall2000}.  
For rigid non-moving targets, advanced registration strategies can yield good compounding results, given that the acquisition protocol is well defined. For example, 
\cite{Wachinger2007} uses defined sweeps and multivariate similarity measures in a maximum likelihood framework to mitigate the problem of registration drift observed in earlier, pair-wise registration methods~\cite{Gee2003}.
However, algorithms requiring all the available images simultaneously to estimate transformations cannot be used in real-time applications such as a visual guidance system for non-expert sonographers to receive feedback, during scanning, of the regions already captured. 

Recent advances in the robustness of semantic discrimination of tissues in medical images largely enabled by the advent of deep learning, and in SLAM algorithms, provide potential to combine these processes in a reliable fashion. 
SLAM is known from natural image processing as a powerful tool for indoor \cite{Whelan2015} and outdoor \cite{Heng2015} mapping, location awareness of robots \cite{Durrant-Whyte2006} and real-time 3D mesh reconstruction from a stream of RGB images that additionally provide depth information \cite{Newcombe2011}. These techniques have been applied in the medical image analysis community to laparoscopy~\cite{Wygant2008} and movement-based diagnosis~\cite{Kontschieder2014}, but never went beyond RGB (+depth) imaging.  

Traditional SLAM methods assume a clear line of sight to map the depth of a scene. However, US images require preprocessing such as segmentation to extract depth of the desired target objects. 
Convolutional neural networks constitute the state of the art for solving (medical) image segmentation tasks eg. \cite{Kamnitsas2017} and have recently shown to be robust for the use in, e.g., fetal screening examinations \cite{Yang2017}, however only at very young GA when the fetus is fully visible in 3D US volumes. 
Our work combines fast automatic tissue segmentation that works also on partially visible tissue in later gestation with modern SLAM algorithms. To the best of our knowledge, this is the first time such an approach is proposed.  

\section{Method}

\begin{figure}[htb]
\centering
\includegraphics[width=\textwidth]{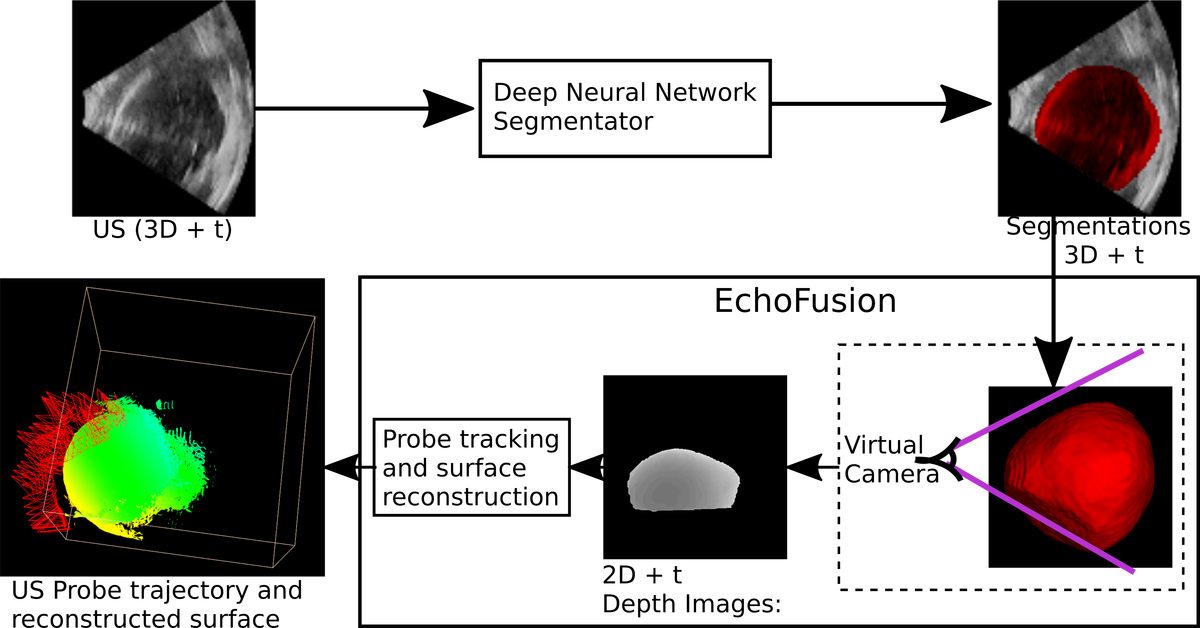}
\caption{Overview: Residual 3D U-Net segments each incoming 3D US from which target fetal organ's surface depth is extracted by a virtual camera located at the ultrasound probe. EchoFusion estimates the camera transformation w.r.t previous frame using the incoming depth image and updates the dense surface model.}
\label{fig:pipeline}
\end{figure}
Our approach consists of three main components: \textbf{(1)} semantic tissue segmentation, 
\textbf{(2)} transducer to object depth map generation, and \textbf{(3)} simultaneous localisation and mapping algorithm. An overview of our approach is shown in Fig.~\ref{fig:pipeline}.

\begin{figure}
\centering
\includegraphics[width=\textwidth]{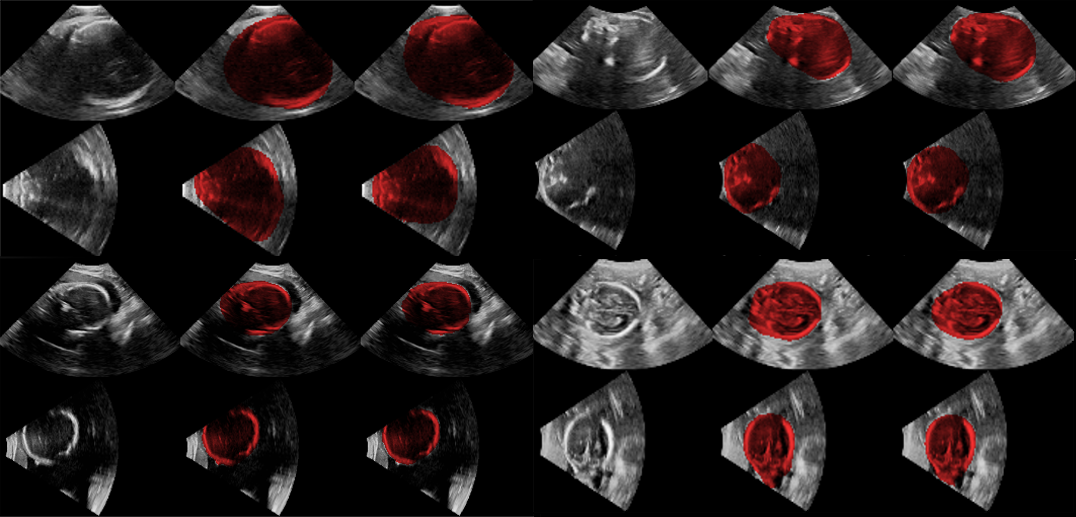}
\caption{Four US volumes with input, GT, and predicted volumes (left to right) of two central orthogonal slices. This shows the typical  diversity of the input sizes, view direction, partial head views, shadows and US artifacts in the dataset used in our experiments.
}
\label{fig:seg_examples}
\end{figure}
\noindent\textbf{(1) Semantic tissue discrimination: } The objective is to produce a binary segmentation of the target object.
For example, for fetal head tracking and reconstruction, the foreground is the fetal head and the remaining structures such as fetal limbs and maternal tissues are background.
Fetal segmentation from freehand 4D US can be quite challenging because of the diversity in the image appearance of the same anatomy, cropping due to limited field of view, and the relatively low quality of 4D images compared to 2D images or static 3D volumes.
As the images are often corrupted by shadows, fetal body surface at distances far from the transducer cannot be delineated as accurately as surfaces physically nearer to the probe.
Thus, in the present work, expert sonographers delineated the closest surface accurately but approximated the shape of the RoI in the surface further from the probe as shown in Fig.~\ref{fig:seg_examples}.

For semantic segmentation, we use a Residual 3D U-Net architecture which has U-Net structure~\cite{Ronneberger2015} and is similar to V-net \cite{milletari2016v} with all convolution layers being replaced by residual-units~\cite{he2016identity} known to make training more stable.
We follow the common strategy whereby skip connections are implemented via concatenation in the up-sampling component of the network and down-sampling is performed with strided convolution (\emph{cf.} max pooling). 
Each convolutional layer of the original architecture is additionally augmented by a residual block containing two convolutions in a similar fashion to~\cite{Kamnitsas2017}. We employ $[16, 32, 64, 128]$ feature maps per layer and all kernels and feature maps are 3D. Each layer additionally utilizes batch normalization, ReLUs and zero-padding.

For training we draw input training patches of size $64\times64\times64$ voxels with an equal probability of patches being centered around a voxel from the foreground or background label class. We train to minimize a standard cross-entropy loss using Adam optimization with learning rate of $0.001$ and $l_{2}$ regularization. Our training imagery is augmented via Gaussian additive noise ($\sigma=0.02$) with image flipping in each axis.

\noindent\textbf{(2) Transducer distance field generation: } Depth images can be generated using a virtual pinhole camera that looks into the 3D segmented model from the same direction as the US probe. All voxels in the output segmentation have known physical co-ordinates with respect to an arbitrary reference point, set as the origin of the world co-ordinate. In the input image volumes, the origin was set to a central point in xz-plane at y=0 making the US probe directed towards positive y-direction and placed $y < 0$. We set a virtual camera that looks towards positive y-axis and along the line $x=z=0$. The exact position and the view angle of the camera depends on the sector width and sector height of the input 3D US volume. 
If the camera is too far away, it sees the flat surface at the edge of the US sector.
Similarly, if the camera is too close, the FoV is not wide enough and some parts of the tissue region may be missed.
In order to estimate an optimal camera position, first we separately compute the intersection and angle between sector lines for the central slices in yz-plane and the central slices in xy-plane as follows:

\vspace{0.1cm}
\noindent\fbox{%
    \parbox{\textwidth}{
\noindent 1. Extract sector mask using thresholding, morphological closing to remove holes.

\noindent 2. Extract edges using Canny edge detection on the sector mask.

\noindent 3. Use Hough transform to detect the two sector lines.

\noindent 4. Compute intersection and angle between the lines found in 3.
}
}
\vspace{0.1cm}

Then, the camera distance is set to be the minimum of the two intersection points, and the view angle is chosen to be the wider of the two angles.

\noindent\textbf{(3) Tracking and Reconstruction with EchoFusion: }  In SLAM~\cite{Newcombe2011}, a sequence of partial views of a 3D scene captured as 2D RGB images and/or depth images is used to estimate all the relative poses of the camera and reconstruct the 3D scene.
Like all SLAM algorithms, we also use only the frontal surface of the 3D scene that are not occluded from the camera view to track and build the 3D scene incrementally.
Thus, we use a volumetric surface representation to store global 3D scene as a truncated signed distance function (TSDF) \cite{Newcombe2011} in a predetermined 3D voxel grid.
This 3D model is updated with each new incoming depth image by estimating the camera transformation with respect to the previous frame. The algorithm can be outlined as follows:

\vspace{0.1cm}
\noindent\fbox{%
    \parbox{\textwidth}{
\noindent 1. From the generated depth image compute the 3D vertex and normals in camera co-ordinate space.

\noindent 2. The 3D vertex and normals from the previous frame are estimated by ray casting the 3D model built so far from the the global camera position estimated from the previous frame.

\noindent 3. The relative camera transformation is then estimated using Iterative Closest Point (ICP) of the two point sets from the current and the previous frames. 
} }
\vspace{0.1cm}

The 3D model gets better and smoother as more consistent data becomes available.


\noindent\textbf{Implementation Details: }
We adapted an open source implementation\footnote{\url{https://github.com/Nerei/kinfu_remake}} of Kinect Fusion~\cite{Newcombe2011}. 
The focal length of the virtual camera can be computed as 
$f = \frac{w/2}{\tan(\alpha / 2)}$, where $\alpha$ is the view angle and $w$ is the image width in pixel co-ordinates.
We set depth and RGB image sizes to $480 \times 480$.
The discriminator model is trained on a Nvidia Titan X GPU with 12 GB of memory.
During runtime, the same GPU can be used for inference and EchoFusion, as the inference from the network does not require large resources like in training time.
The network was implemented in tensorflow.

\section{Experiments and Results}

\noindent\textbf{Phantom data: }
We use data from a fetal phantom Kyotokagaku UTU-1 at a gestational age of about 20 weeks.
The GT segmentation consists of fetal vs. maternal tissue delineation in 28 3D volumes which is randomly split into 24 training samples and 4 validation samples.
The GT segmentations include both the fetal head and body as foreground.

\noindent\textbf{Fetal screening data: } 
Two expert sonographers delineated 192 US fetal head volumes  for training and validation of fetal head segmentation. 
These 3D images were selected from 4D freehand scanning of 19 different fetuses having GAs in the range of 23-34 weeks with mean (std) age of 30 (2.842) weeks.

The sonographers used MITK~\cite{Wolf_themedical} to segment six to seven representative slices manually, then performed 3D interpolation from these slices to create a 3D shape.
Many of these images contained shadows on the far-field surface, so the manual delineation was done empirically based on the sonographers' anatomical knowledge of the head shape. 
We split 192 GT data into 184 training and 8 validation images.
We then test the trained network only once on a set containing GT segmentations from five fetuses not used in training-validation set.  
\noindent\textbf{Evaluation: } We use Dice score to evaluate the performance of segmentation quantitatively. Evaluating tracking accuracy is challenging without a ground truth. Surface reconstruction which can be qualitatively observed depends on the tracking obtained from the SLAM. To assess the tracking robustness on freehand 4D US stream of the real fetal heads, we test our framework on 37 fetuses and compute the number of tracking losses (i.e. reset of the tracked pose) and the longest sequence without any resets.

\begin{figure}
\centering
\subfloat[]{\includegraphics[height=4cm]{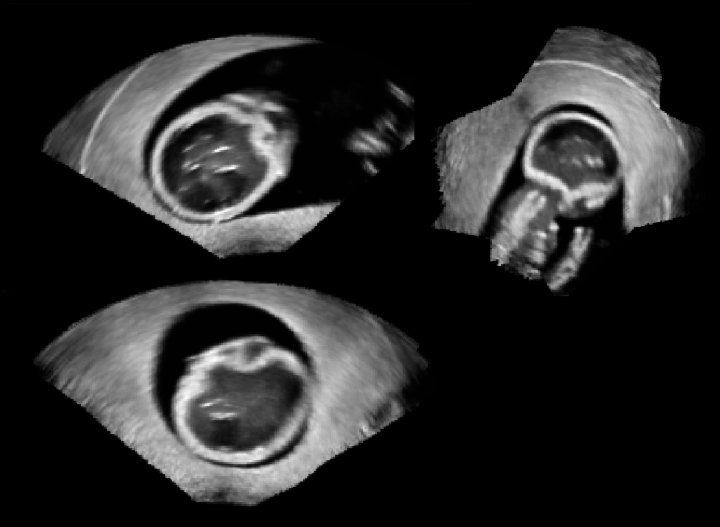}}
\subfloat[]{\includegraphics[height=4cm]{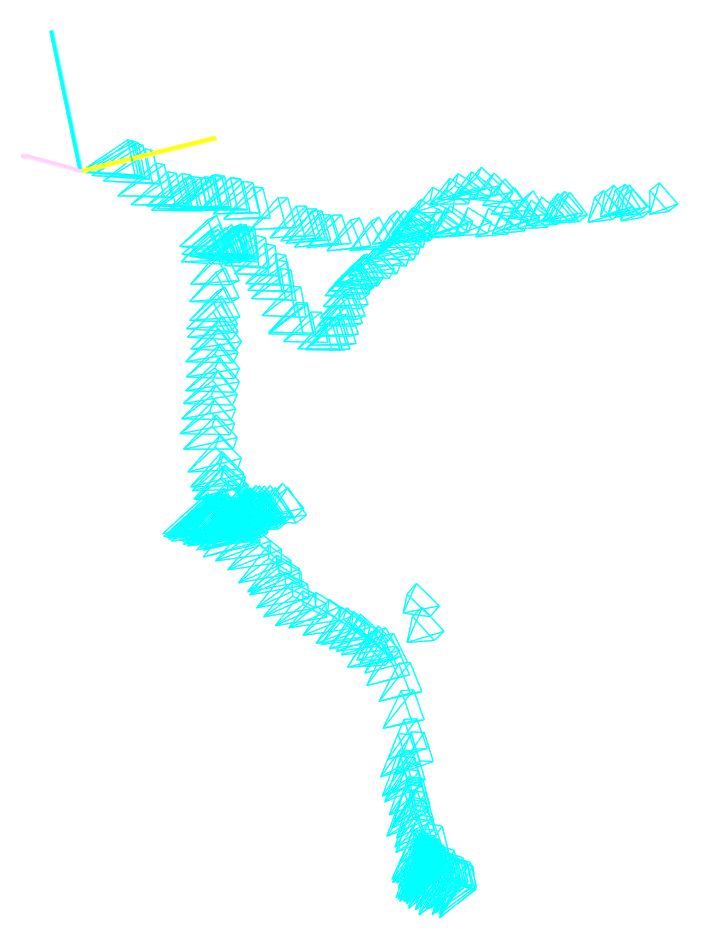}}
\subfloat[]{\includegraphics[height=4cm]{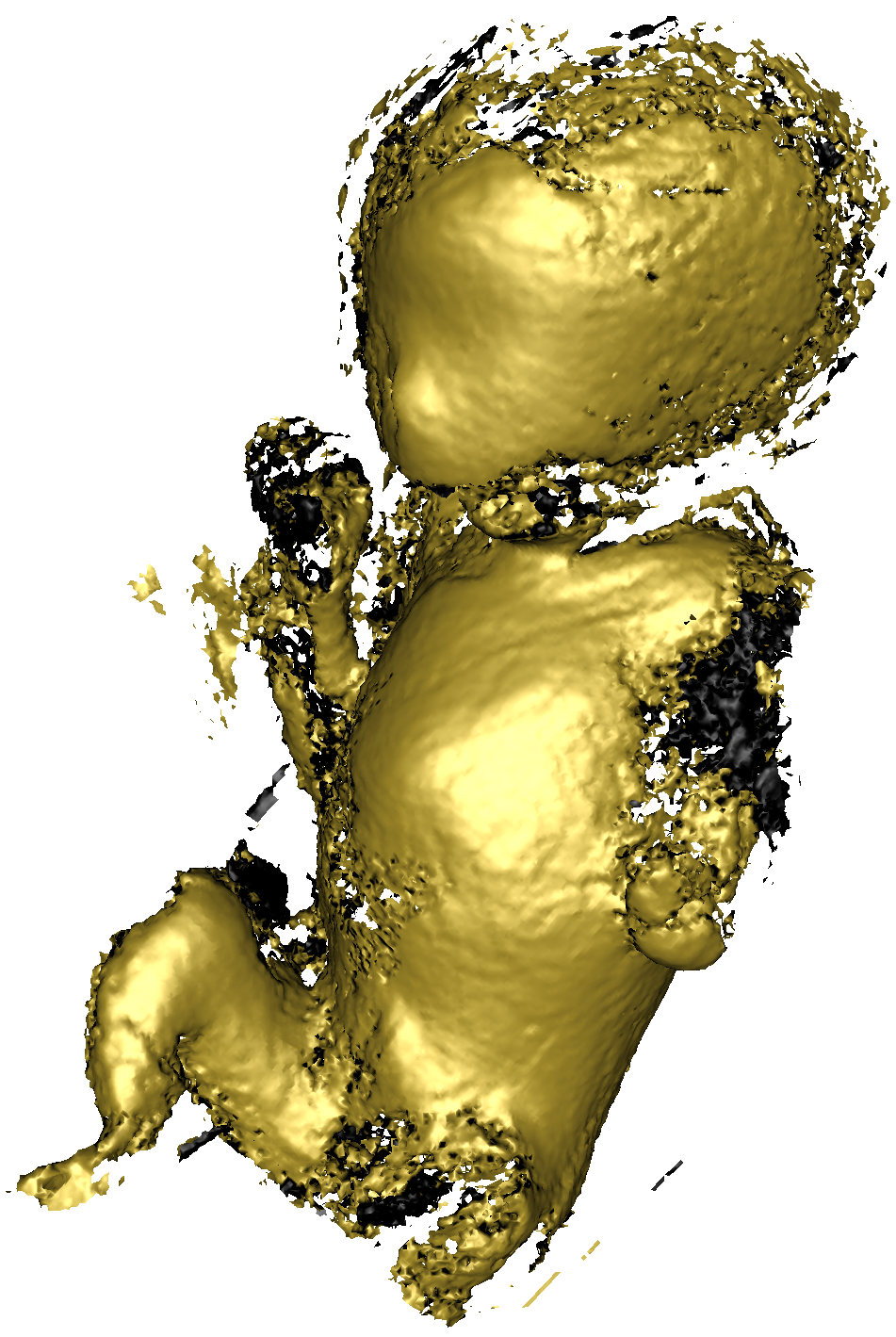}}
\caption{Orthogonal slices through examples of compounded volumes (a), EchoFusion tracking trajectory (b) and TSDF iso-surface reconstruction (c) for sequences from whole body fetus phantom. The sequence of images of the static phantom were taken with a very wide range of probe directions as seen in the top right slice in (a), and from the trajectory in (b). Limbs are not reconstructed faithfully due to limb information being purposefully discarded at segmentation time.}
\label{fig:resultsPhantom}
\end{figure}

\begin{figure}
\centering
\subfloat[]{\includegraphics[height=4cm]{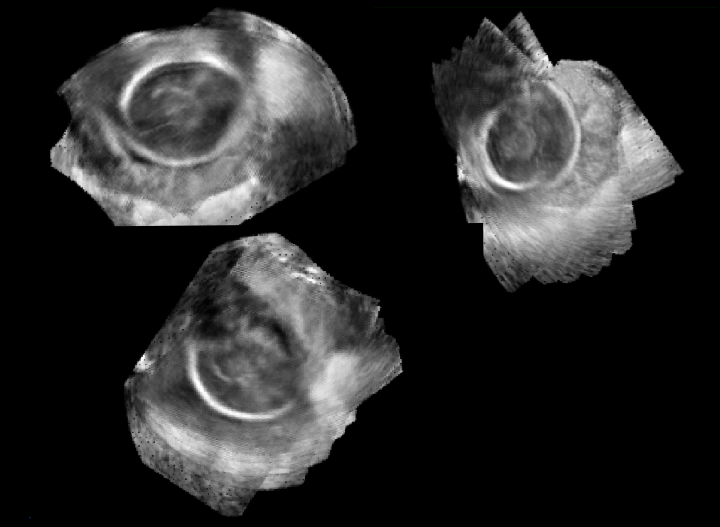}}
\subfloat[]{\includegraphics[height=4cm]{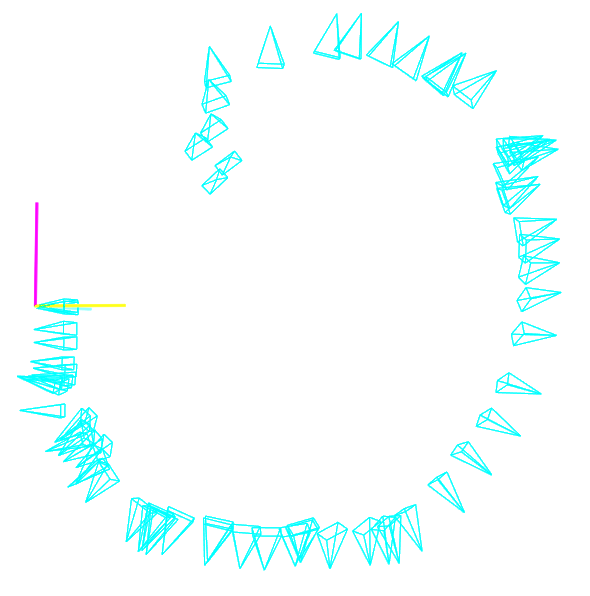}}
\subfloat[]{\includegraphics[height=4cm]{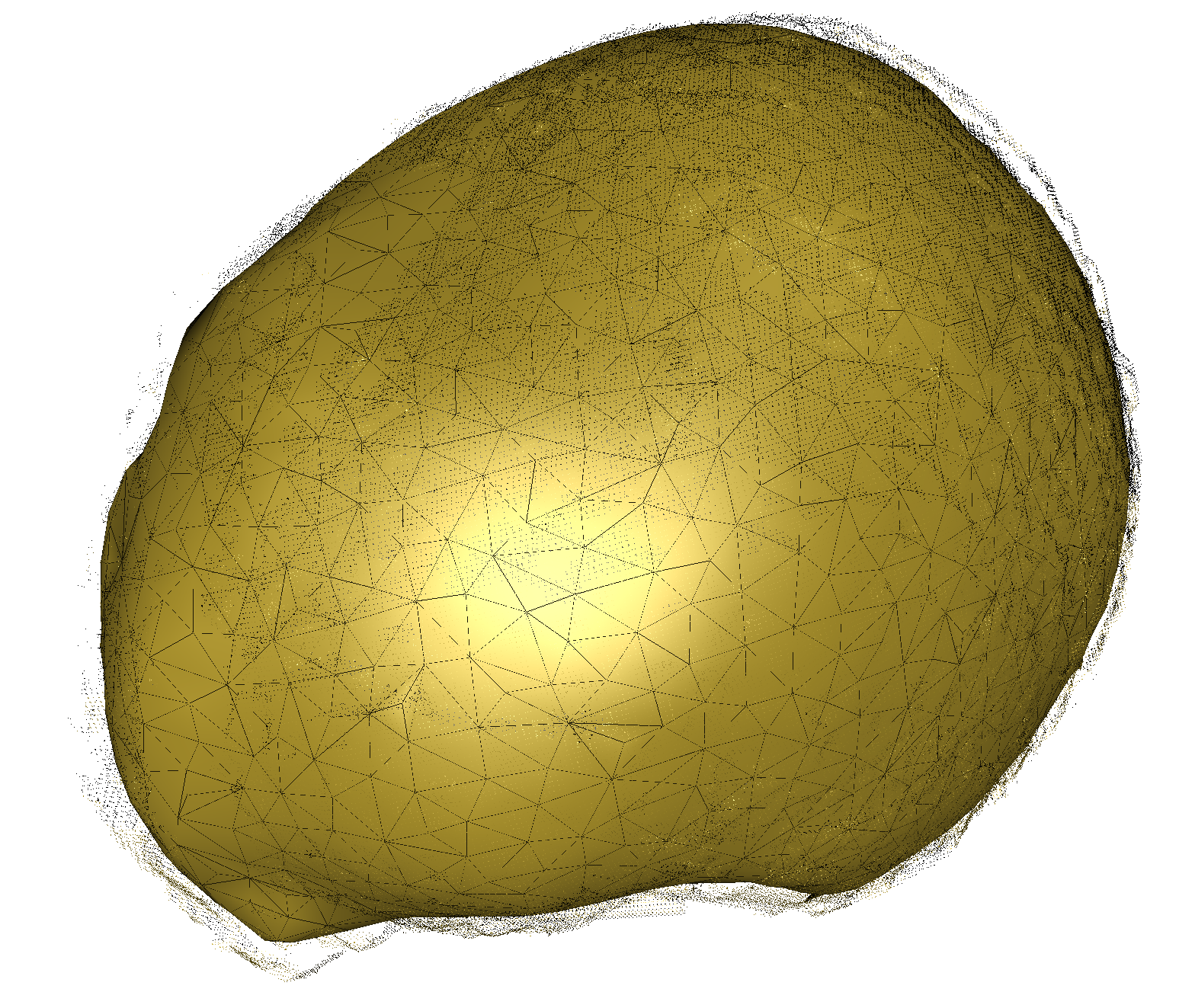}}
\caption{Orthogonal slices through examples of compounded volumes (a), EchoFusion tracking trajectory (b) and TSDF iso-surface reconstruction (c) for sequences of a real fetal head. Note that the tracking is relative only to the fetal head, and not the other moving maternal and fetal tissues.}
\label{fig:resultsReal}
\end{figure}

\noindent\textbf{Results: }
Table~\ref{tab:dice} shows quantitative results for segmentation performance on both the phantom and the real fetuses.
Since there was only one phantom available which was used to create training and validation set, there is no test set for the phantom.
For the real fetuses, test set was created using the same protocol as the training sets but from the fetuses that were not used for training or validation. Although the number of images used for training on the phantom is much smaller than for the real fetuses, the validation set accuracy is higher for the phantom. This is not surprising because the images from the phantom are much less challenging than the real fetuses.

Fig.~\ref{fig:resultsPhantom} and Fig.~\ref{fig:resultsReal} show qualitative results after compounding a series of 10-20 EchoFusion-tracked consecutive 3D volume acquisitions from different locations.
3D surface reconstruction in Fig.~\ref{fig:resultsPhantom} shows that both the phantom face and body which were selected as foreground objects for the segmentation are nicely reconstructed.  
Similarly, the fetal head in compounded volume in Fig.~\ref{fig:resultsReal} shows that the sequence of images registered reasonably well although they were taken from a wide range of angles.    
Table~\ref{tab:track} shows EchoFusion tracking performance on 37 fetal sequences of volumes.
On average, there were approximately 98 total frames for which the SLAM algorithm lost tracking approximately 5 times.
These sequences were obtained by moving the probe in different directions trying to cover the head (skull and face) from all possible directions.
The sequences were used as they were acquired without data cleaning, thus containing views which do not show the fetal head and many frames with only partial views of the head region.

\vspace{-2mm}
\begin{table}[ht]
\centering
\caption{Dice scores for real-time semantic tissue discrimination.}
\label{tab:dice}
\begin{tabular}{lcccc}
\toprule
set & images(real) & ~~~mean(std)~~~ & images(Phantom) & ~~~mean(std~~~\\
\midrule
Train & 178  & 0.9408(0.0389) & 24 & 0.9735(0.0125)\\
Validation & 8 & 0.9217(0.0212) & 4 & 0.9267(0.0074)\\
Test  & 26 & 0.8942(0.0671) & - & - \\
\bottomrule
\end{tabular}
\end{table}

\begin{table}[ht]
\centering
\caption{Robustness with respect to continuously tracked frames for 37 fetuses.
}
\label{tab:track}
\begin{tabular}{lccc}
\toprule
 ~~~~ ~~      & mean (std) & median & range \\
\midrule
total frames & 98.11(54.65) & 91 & [21, 277] \\
no. of tracking losses & 5.16(3.67) & 5 & [0, 15] \\
longest sequence without tracking loss & 40.86(30.85) & 31 & [4, 152] \\
\bottomrule
\end{tabular}
\end{table}

\section{Discussion}
The key contribution of this work is the novel approach to the tracking and compounding problem in freehand 4D US, which constitutes combining the powerful semantic segmentation neural networks with modern SLAM algorithms.
Since both of them are very active fields of research, there is a lot of potential to improve EchoFusion for a multitude of applications including compounding, image reconstruction, artefact reduction, super resolution and fetal face biometrics using the resulting dense surface model. Moreover, this method could also allow non-expert to acquire dense data for retrospective evaluation. 

The goal of this work was to provide a proof of concept, but clinical translation of this method would require a more extensive quantitative validation of the tracking accuracy, drift over the long sequence, and compare how segmentation accuracy impacts the overall tracking accuracy.

The use of whole body phantom vs fetal head also demonstrates that the top level approach generalises across organs and anatomy as we can train the segmentation network for a desired RoI.
However, the current implementation of the SLAM algorithm works only for largely rigid body motion; the static phantom and the fetal head can be reasonably assumed to have mostly rigid body movement with respect to the probe at the semantic level. 
For non-rigid movements of the fetus such as the whole body or abdomen, the current SLAM component must be replaced with the methods that take dynamic scene changes into account~\cite{slavcheva2018sobolevfusion}.
However, such approaches would still not be robust to sudden movements (\emph{e.g.} kicks) and introduce a significant computational overhead, potentially jeopardizing hard real-time constraints. One approach to tackle this problem is to consider such suddenly moving limbs as background in segmentation so that they are ignored during the tracking and reconstruction. There can still be challenges, (\emph{e.g.} turning the head in the opposite direction and staying there, when reconstructing head/shoulders/torso at once), and is more of an open problem at present. However, being able to focus and compound on quasi-rigid areas like only the head or only abdomen and changing the model depending on target application would already be very valuable e.g., for the creation of fetal brain or abdomen atlases.

\section{Conclusion}
We have developed a novel approach demonstrating a promising potential for robust segmentation and tracking of fetuses in utero. EchoFusion is versatile and could be applied in any situation where an independently moving target object is occluded by other tissue or material. We have also introduced a way to learn a tissue discriminator from weak annotations in fetal 3D US images and discussed the performance of a Residual 3D U-Net tissue discriminator learning from this data. This discriminator is key to establishing semantics for SLAM-based tracking, which we evaluated on 4D freehand US of a fetal phantom and on real fetuses from screening examinations. In the future, we will perform a more extensive validation of the tracking accuracy, and also find a way to derive robust SDFs from tissue probabilities to exploit the possibilities of dynamic fusion approaches.   

\noindent\textbf{Acknowledgements:} This work was supported by the Wellcome/EPSRC Centre for Medical Engineering [WT 203148/Z/16/Z], Wellcome Trust IEH Award [102431]. The authors thank Nvidia Corporation for the donation of a Titan Xp GPU.
\bibliographystyle{splncs03}
\bibliography{references}

\end{document}